\documentclass[]{style/interact}

\usepackage{epstopdf}

\usepackage{subfigure}
\usepackage{amsmath}
\usepackage{autobreak}
\usepackage{bm}
\usepackage{enumitem}

\usepackage[numbers,sort&compress]{natbib}
\bibpunct[, ]{[}{]}{,}{n}{,}{,}
\renewcommand\bibfont{\fontsize{10}{12}\selectfont}
\makeatletter
\def\NAT@def@citea{\def\@citea{\NAT@separator}}
\makeatother

\theoremstyle{plain}

\theoremstyle{definition}

\theoremstyle{remark}

\usepackage[markup=default, defaultcolor=red]{changes}
\def\revised_color{red}

\begin{document}

\articletype{ADVANCED ROBOTICS}

\title{Design Methodology of Hydraulically-driven Soft Robotic Gripper for a Large and Heavy Object}

\author{
Ko Yamamoto\textsuperscript{a} $^{*1}$\thanks{$^{*1}$CONTACT Ko Yamamoto. Email: yamamoto.ko@ynl.t.u-tokyo.ac.jp}, \name{Kyosuke Ishibashi\textsuperscript{a}, Hiroki Ishikawa\textsuperscript{a} and Osamu Azami\textsuperscript{a}}
\affil{\textsuperscript{a}Department of Mechano-informatics, The University of Tokyo 7-3-1 Hongo, Bunkyo-ku, Tokyo 113-8656, Japan}
}

\maketitle

\begin{abstract}
This paper presents a design methodology of a hydraulically-driven soft robotic gripper for grasping a large and heavy object --- approximately 10 - 20 kg with 20 - 30 cm diameter. Most existing soft grippers are pneumatically actuated with several hundred kPa pressure, and cannot generate output force sufficient for such a large and heavy object. Instead of pneumatic actuation, hydraulic actuation has a potential to generate much larger power by several MPa pressure.
In this study, we develop a hydraulically-driven soft gripper, in which its basic design parameters are determined based on a mathematical model that represents the relationship among the driving pressure, bending angle, object mass and grasping force. 
Moreover, we selected materials suitable for grasping a heavier object, based on the finite element analysis result of the detailed design. We report experimental results on a 20 kg object grasping and closed-loop control of the finger bending angle. 
\end{abstract}

\begin{keywords}
Soft Robotic Gripper; Hydraulics; Modeling and Control for Soft Robots
\end{keywords}

\renewcommand{\thefootnote}{\fnsymbol{footnote}}
\footnote[0]{\copyright \: 2026 Copyright held by the owner/author(s). This is an original manuscript of an article published by Taylor \& Francis in Advanced Robotics on 8th Jan. 2026, available at: https://doi.org/10.1080/01691864.2025.2608943}
\renewcommand{\thefootnote}{\arabic{footnote}}

\section{Introduction}

Soft robotic grippers have gained significant attention due to their adaptability to various objects \cite{shintake2018soft, Rus2015} and have a lot of applications including food industry, agriculture, healthcare, manufacturing, logistics \cite{Fantoni2014Grasping, zhang2020state, wang2022challenges}. 
Especially, soft grippers are suitable for a lightweight, delicate and fragile object with irregular shape, which traditional rigid robotic grippers cannot grasp safely and effectively. Several types of soft grippers have been developed, each utilizing different materials, actuation methods, and design principles, such as pneumatically actuated grippers \cite{xavier2022soft}, electroactive polymer grippers \cite{gupta2019soft, park2022recent}, shape memory alloy (SMA) grippers \cite{xia2021review}, magnetorheological (MR) fluid grippers \cite{pettersson2010design}, electroadhesion grippers \cite{guo2019electroadhesion}, gripper with Gecko-inspired adhesives \cite{ruotolo2021grasping}, soft grippers with jamming mechanisms \cite{amend2016soft} and Origami-inspired grippers \cite{li2019vacuum}. 

However, most existing soft grippers are pneumatically actuated with several hundred kPa pressure, and cannot generate output force sufficient for such a large and heavy object that usually human is capable of --- approximately 10 - 20 kg with 20 - 30 cm diameter.
According to the Health and Safety Executive in the UK \cite{HSE}, a human worker is encouraged to carry an object less than 16 kg for woman and 25 kg for man to reduce the risk of injury. 
There are a few attempts to develop a soft gripper that can grasp a heavy object. Yap et al. \cite{Yap2016} developed a soft robot gripper capable of grasping objects up to 10 cm and 5 kg. However, it was reported \cite{Yap2016} that its output power decreases as the target object size increases. Hwang et al. \cite{hwang2021electroadhesion} developed a gripper using electro-adhesion that was capable of 16.8 kg. However, this gripper was not suitable for an object with rough and moist surfaces.

Instead of pneumatic actuation, hydraulic actuation has a potential to generate much larger power by several MPa pressure. 
Previously, Hagiwara et al. \cite{Hagiwara2022} developed a hydraulically-driven soft gripper capable of grasping a 5 kg object. 
Then, we developed a prototype of a hydraulically-driven soft gripper for transporting large and heavy vegetables in a vegetable factory \cite{Azami2022,Ishibashi2023_ar}. 
However, those studies focused on the design of the soft finger with durability of high pressure, and the analysis on the actuation performance by a hydraulic pump.
In the design and control of a soft gripper, it is important to analyze the total system mechanics consisting of hydraulic pump, soft fingers and target object.
In \cite{ishibashi2024}, we constructed a mathematical model of the soft gripper including the actuation unit, and showed that the oil reservoir with rubber sheet can boost the pump actuation and decrease the pressure necessary for bending the soft fingers.

In this paper, we present a design methodology of a hydraulically-driven soft robotic gripper for grasping an object with 10 to 20 kg weight and 20 to 30 cm diameter. 
We estimate the maximum payload based on the previously developed model and the friction condition. 
Then, we conduct a strength analysis by finite element method (FEM) in the detailed design of the soft gripper and select appropriate material for grasping a 20 kg object.
Finally, we report experimental results on the 20 kg payload test, grasping various objects, and a closed-loop control based on the proposed model.

The rest of this paper is organized as follows. Section \ref{chap:modeling} presents the basic structure of the soft gripper and its mathematical model. Section \ref{chap:soft_hand} details the design based on the parameters obtained in Section \ref{chap:modeling} and provides the FEM analysis. Section \ref{chap:evaluation} reports the experimental validations. 
Finally, Section \ref{sect:discussion} provides discussions on the obtained results, and Section \ref{chap:conclusion} concludes this paper.

Portions of this paper were previously presented in \cite{ishibashi2024}, including the basic of mathematical model (Section \ref{chap:modeling}.2), and the structure of the soft gripper (Section \ref{chap:soft_hand}.1). 
Whereas the previous paper \cite{ishibashi2024} focused on the feedback control of the bending
angle and the evaluation of the grasping flexibility achieved by the feedback, the main contribution of this study is to present a design methodology to achieve high payload such as 20 kg and to provide its experimental validations.
\section{Design Methodology of Hydraulically-driven Soft Robot Gripper}
\label{chap:modeling}

\subsection{Structure of Hydraulically-driven Soft Robot Gripper}

\begin{figure}[t]
    \centering
    \subfigure[Actuation system for soft robot gripper]{
        \includegraphics[width=0.45\hsize]{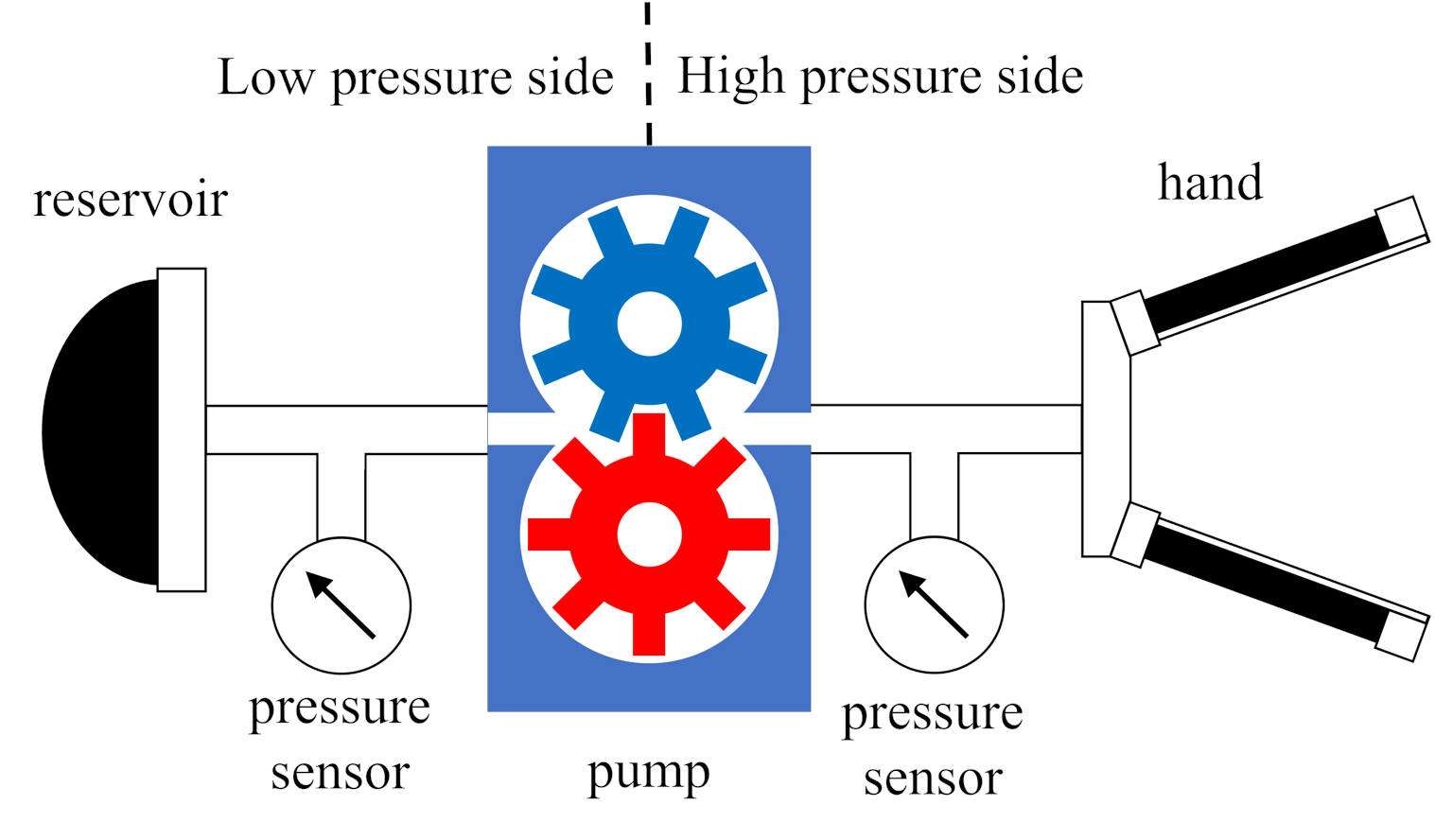}}
    \subfigure[Soft finger]{
        \includegraphics[width=0.26\hsize]
        {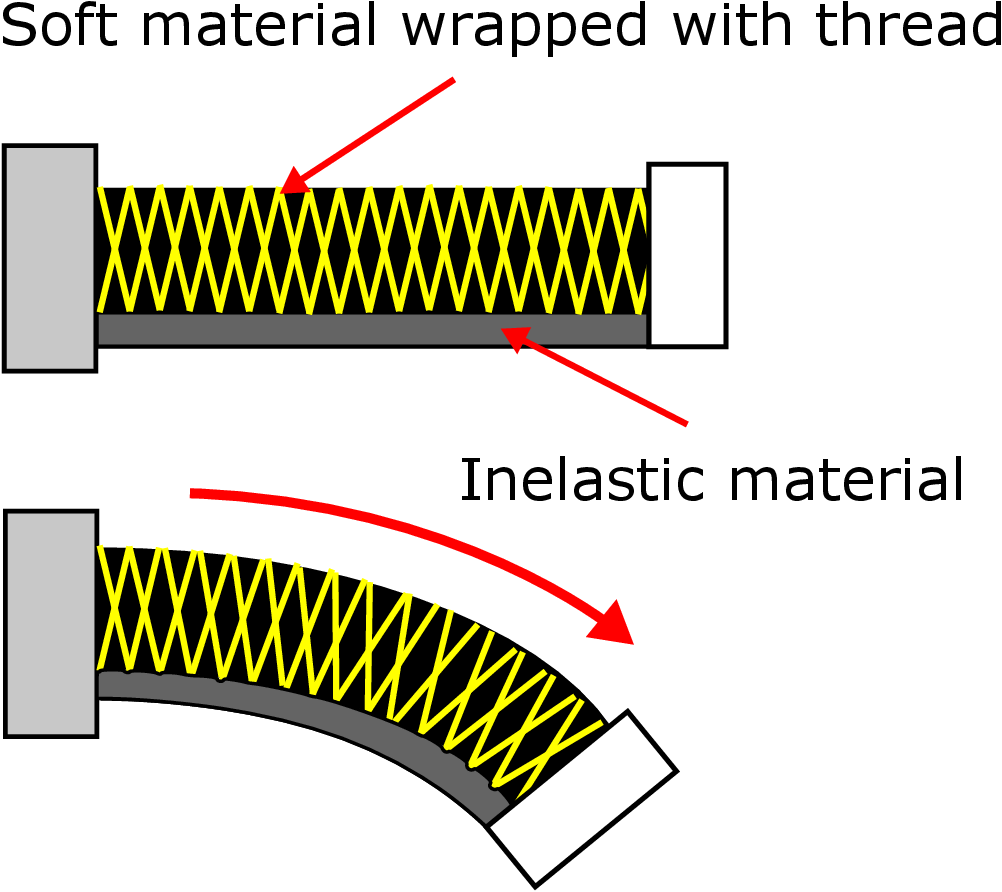}}
    \subfigure[Reservoir to store oil]{
        \includegraphics[width=0.23\hsize]{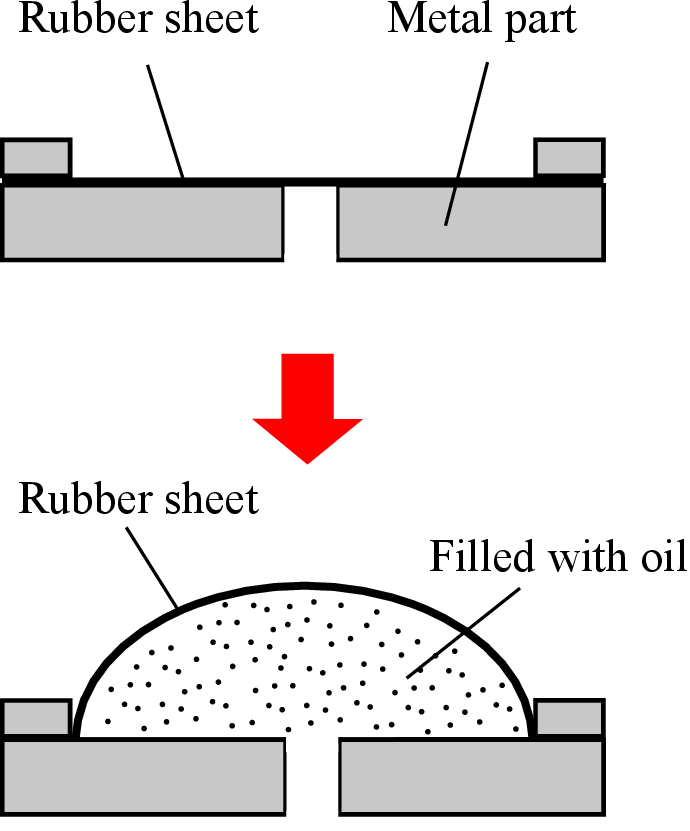}}
    \caption{(a) Soft fingers are connected to the high-pressure side of the hydraulic pump while a reservoir is connected to the low-pressure side. The pump generates differential pressure, sending oil from the reservoir to the soft fingers. (b) Structure and bending mechanism of fiber-constrained soft finger. Fiber-constrained soft fingers utilize the expansion of soft materials under pressure and deform only in a certain direction by embedding inelastic materials. (c) Structure of reservoir. The reservoir consists of a circular sheet made of NBR, along with a ring-shaped metal part and base.}
    \label{fig:simple_diagram}
\end{figure}

Fig. \ref{fig:simple_diagram} shows the structure of the hydraulically-driven soft robot gripper including its actuation system. 
A hydraulic pump generates a differential pressure, in which the high- and low-pressure sides are connected to soft fingers and oil reservoir, respectively.
The differential pressure sends the oil from the reservoir to the soft fingers and pressurizes the inner flow path of the soft finger. 
We employ the fiber-constrained design \cite{Suzumori1991, deimel2016novel}, in which the radial expansion is constrained by wrapping a thread around the finger, as shown in Fig. \ref{fig:simple_diagram} (b). 
When the inner flow path is pressurized, the finger expands only in the axial direction. 
This expansion is transformed into a bending deformation because one side of the finger is constrained by an inelastic part. 
Fig. \ref{fig:simple_diagram} (c) illustrates the structure of the oil reservoir, consisting of a circular sheet made of NBR and a ring-shaped metal part and base.

\subsection{Modeling of Hydraulically-driven Soft Robot Gripper \cite{ishibashi2024}}

\begin{figure}[t]
    \centering
    \includegraphics[width=0.7\hsize]{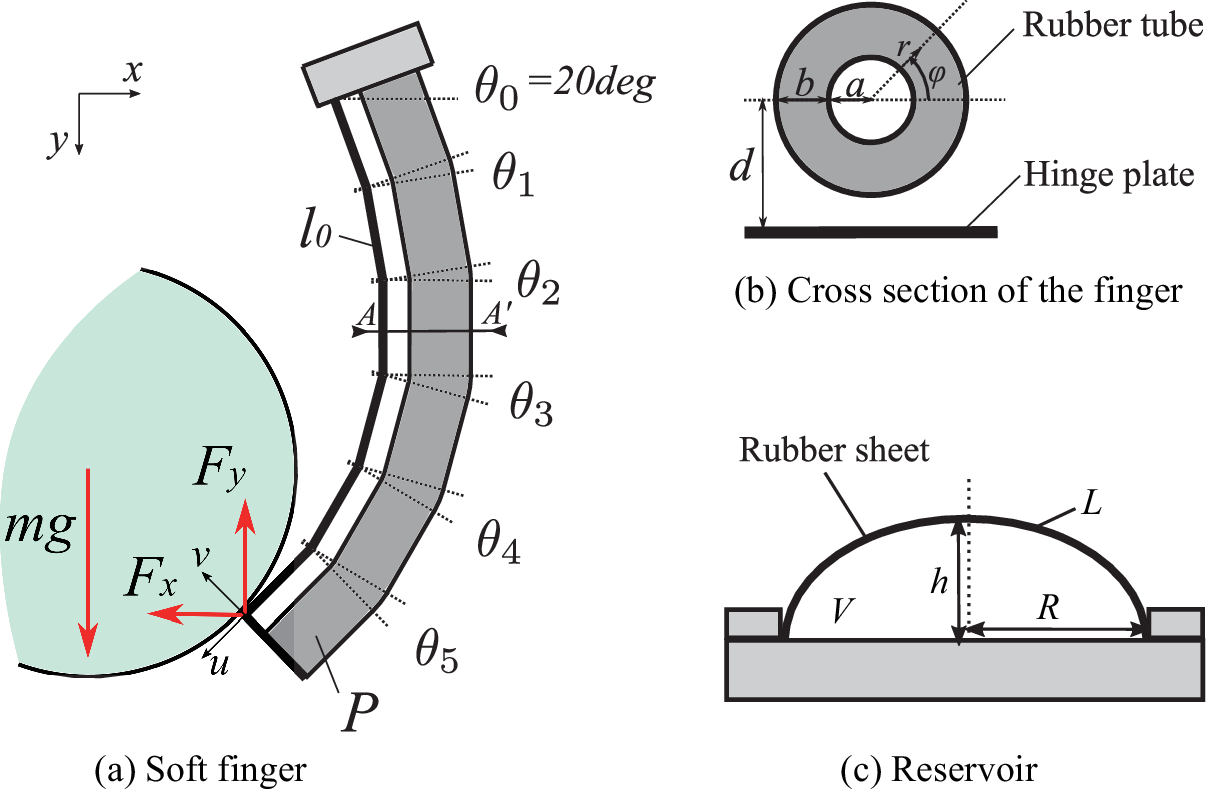}
    \caption{Geometric Model of hydraulic soft gripper. (a) Soft finger. (b) Cross section of finger. (c) Reservoir made of rubber sheet. Reservoir consists of a circular sheet made of NBR, along with a ring-shaped metal part and base, as shown in Fig. \ref{fig:simple_diagram} (b).}
    \label{fig:finger_model}
\end{figure}

We formulated a mathematical model to represent the relationship among the driving pressure, bending angle of a finger, and grasping force on a target object. Renda et al. \cite{renda2018, renda2016discrete} proposed a PCS model that divides a flexible rod structure into a finite number of segments with constant strain, deriving its kinematics and dynamics. Polygerinos et al. \cite{polygerinos2015, wang2016} modeled fiber-reinforced soft fingers assuming constant curvature, while Mustaza et al. \cite{Mustaza2019} developed a manipulator with fiber-constrained actuators and used Lagrange's equations for dynamic modeling. Sedal et al. \cite{sedal2018continuum} derived equations for fiber-constrained soft actuators, relating various mechanical parameters under static conditions using continuum mechanics. However, these studies focused on modeling the soft finger alone, without considering the entire gripper system. In \cite{ishibashi2024}, we proposed a mathematical model that includes both the soft fingers and the reservoir and demonstrate that the reservoir's elasticity contributes to efficient actuation.

\begin{table}[t]
    \centering
    \footnotesize
    \begin{minipage}{1.0\hsize}
    \caption{Parameters and Symbols of the gripper model.}
    \label{tb:model_param}
    \centering
    \begin{tabular}{c|c|c}
        Param & Description & Value \rule[-5pt]{0pt}{10pt} \\ \hline
        $\theta_0$ & Inclination of the base link of the finger & 20deg \rule[-7pt]{0pt}{16pt} \\
        $l_0$ & Initial length of the finger & 180mm \rule[-7pt]{0pt}{14pt} \\
        $a$ & Internal diameter of the rubber tube  & 4mm \rule[-7pt]{0pt}{14pt} \\
        $b$ & Thickness of the rubber tube & 5mm \rule[-7pt]{0pt}{14pt} \\
        $d$ & Distance between hinge plate and rubber tube & 15mm \rule[-7pt]{0pt}{14pt} \\
        $R$ & Radius of reservoir sheet & 24mm \rule[-7pt]{0pt}{14pt} \\
        $t$ & Thickness of reservoir sheet & 1mm \rule[-7pt]{0pt}{14pt} \\
        $V_0$ & Initial volume of oil stored in reservoir & 17.8$\rm{cm^3}$ \rule[-7pt]{0pt}{14pt} \\
        $n$ & The number of finger attached to the gripper & 4 \rule[-7pt]{0pt}{14pt} \\
        $n_s$ & The number of hinge joint in a finger & 5 \rule[-7pt]{0pt}{14pt} \\
        $\mu_{sf}$ & Initial shear modulus of rubber tube (finger)& 1.15MPa \rule[-7pt]{0pt}{14pt} \\
        $\mu_{sr}$ & Initial shear modulus of rubber sheet (reservoir) & 1.15MPa \rule[-7pt]{0pt}{14pt} \\
        $\mu_f$ & Coefficient of static friction & 0.7 \rule[-7pt]{0pt}{14pt} \\
    \end{tabular}
    \end{minipage}
    \begin{minipage}{1.0\hsize}
    \centering
    \vspace{5mm}
    \begin{tabular}{c|c}
        Symbol & Description \rule[-5pt]{0pt}{10pt} \\ \hline
        $P$ & Pressure of the inner side of the finger \rule[-7pt]{0pt}{16pt} \\
        $F_x, F_y$ &  Force applied to the tip of the finger \rule[-7pt]{0pt}{14pt} \\
        $(\theta_1, \cdots ,\theta_{n_s})^T$ & Bend angles between the hinge plate \rule[-7pt]{0pt}{14pt} \\
        $\theta = \Sigma \theta_i$ & Bend angle of finger \rule[-7pt]{0pt}{14pt} \\
        $m$ & Mass of the grasping object \rule[-7pt]{0pt}{14pt} \\
    \end{tabular}
    \end{minipage}
\end{table}

Fig. \ref{fig:finger_model} shows a geometric model of a soft finger, and Table \ref{tb:model_param} lists variables used in the proposed model. 
We simplified the finger structure based on the following assumptions:
\begin{itemize}
    \item The soft robot gripper has multiple fingers, but we assume that all fingers have the same deformation for simplicity.
    \item The inelastic part is divided into multiple segments connected by revolute joints.
    \item When the gripper grasps an object, the weight is supported only at the fingertips.
    \item The contact forces are equally distributed among the fingertips.
\end{itemize}

From the virtual work principle, the virtual work by the pressure $P$ and contact force $(F_x, F_y)$ is equal to the variations of the strain energies in the fingers and the reservoir.
This relationship is represented as
\begin{align}
    P \Delta Q + n (F_x \Delta x + F_y \Delta y) = n \Delta W_f + \Delta W_r .
    \label{eq:virtual_work}
\end{align}
Outline of each term is summarized as follows:
\begin{itemize}
    \item The first term on the left-hand side represents the virtual work produced by the hydraulic pump, where $\Delta Q$ denotes the flow rate in the finger. 
    \item The second and third terms account for the work produced by the contact force, where $n$ is the number of fingers, and $\Delta x$ and $\Delta y$ are the virtual displacements of the fingertip position.
    \item In the first term on the right-hand side, $\Delta W_f$ is the variation in strain energy of the finger rubber tube.
    \item $\Delta W_r$ is the variation in strain energy in the reservoir sheet. 
\end{itemize}
The details are summarized in the following subsections.

\subsubsection{Virtual Work by Hydraulic Pump}
The virtual work by the hydraulic pump is given by $P \Delta Q$, where $Q$ is the internal volume of the fingers. 
Let $\theta_i \ (i=0, \cdots, n_s)$ denote the bending angle of the $i$-th segment. 
Assuming a finger has a simple cylindrical shape as shown in Fig. \ref{fig:finger_model}, $Q$ is calculated as
\begin{align}
    Q = n a^2 \pi (l_0 + d \theta)
    \label{eq:volume}
\end{align}
where
\begin{align}
    \theta := \sum_i \theta_i .   
\end{align}
Therefore, the virtual work $P \Delta Q$ is calculated as
\begin{align}
    P \Delta Q =  P n a^2 \pi d \Delta \theta
    \label{eq:pump_work}
\end{align}
where $\Delta \theta$ is the variation of $\theta$.

\subsubsection{Variation in Strain Energy of a Finger}
Next, we consider the variation in strain energy $\Delta W_f$ of the NBR tube.
The strain energy $W_f$ of a finger is calculated based on the Neo-Hookean model \cite{xavier2021finite} as
\begin{align}
    W_f = \int \frac{\mu_{sf}}{2} (I_1-3) dV
    \label{eq:finger_strain_energy}
\end{align}
where $I_1$ is the first invariant of 
deviation strain, represented using the axial, radial and circumferential deformation rates $\lambda_1, \lambda_2, \lambda_3$ as
\begin{align}
     I_1 = \lambda_1^2 + \lambda_2^2 + \lambda_3^2 .
     \label{eq:deviation_distortion}
\end{align}
In this study, $\lambda_2=1$ is satisfied because the radial direction is constrained. 
Moreover, $\lambda_1 \lambda_2 \lambda_3=1$ is satisfied because NBR is an incompressible material. 
Therefore, the values of $\lambda_1, \lambda_2$ and $\lambda_3$ are summarized as
\begin{align}
     \lambda_f := \lambda_1, \: \lambda_2 = 1, \: \lambda_3 = \frac{1}{\lambda_f}
     \label{eq:lambda_if}
\end{align}
where $\lambda_f$ is an alias of $\lambda_1$, 
Considering a polar coordinate $r$ and $\phi$ in the cross section A-A' shown in Fig. \ref{fig:finger_model}, $\lambda_f$ is calculated as
\begin{align}
        \lambda_f = \frac{l_0 + \{ d +(a+r)\sin{\phi} \} \theta}{l_0}
    \label{eq:lambda_f}
\end{align}
From (\ref{eq:finger_strain_energy})-(\ref{eq:lambda_f}), we can represent $\Delta W_f$ as
\begin{align}
        \Delta W_f &= \mu_{sf} \zeta_f(\theta) \Delta \theta
    \label{eq:dw_f}
\end{align}
where
\begin{align}
    \zeta_f(\theta) &:= \int_0^{2 \pi} \int_0^b (\lambda_f - \frac{1}{\lambda_f^3}) \frac{d \lambda_f}{d \theta} \cdot l_0 (a + r) dr d\phi .
    \label{eq:zeta_f}    
\end{align}

\subsubsection{Variation in Strain Energy of Reservoir Sheet}
The strain energy in the reservoir sheet $W_r$ can be expressed as
\begin{align}
    \label{eq:reservoir_strain_energy}
    W_r &= \int \frac{\mu_{sr}}{2} (I_1-3) dV
\end{align}
in a manner similar to \eqref{eq:finger_strain_energy}.
Because the rubber sheet undergoes biaxial tensile deformation, the values of $\lambda_1,\lambda_2$ and $\lambda_3$ are given as
\begin{align}
    \lambda_r := \lambda_1 = \lambda_2, \lambda_3 = \frac{1}{\lambda_r^2}
    \label{eq:lambda_ir}
\end{align}
where $\lambda_r$ is an alias of $\lambda_1$ or $\lambda_2$.

Fig. \ref{fig:finger_model} (c) shows a schematic diagram of the reservoir. Assuming that the rubber sheet deforms into an ellipse, the circumference $L(h)$ of the ellipse can be formulated using Ramanujan's approximation \cite{barnard2001inequalities} as
\begin{align}
    2L(h) = \pi \{ 3(R+h) - \sqrt{(R+3h)(h+3R)} \}
    \label{eq:eclipse}
\end{align}
where $R$ and $h$ are the radius and height of the reservoir, respectively, calculated as
\begin{align}
    h &= \frac{3}{2 \pi R^2} V, \\
    V &= V_0 - a^2 \pi d n \theta
    \label{eq:reservoir_volume}
\end{align}
Then, $\lambda_r$ can be written as
\begin{align}
    \lambda_r = \frac{L(h)}{2R}
    \label{eq:reservoir_lambda}
\end{align}
From (\ref{eq:reservoir_strain_energy})-(\ref{eq:reservoir_volume}), we can calculate $\Delta W_r$ as
\begin{align}
        \Delta W_r = \mu_s \zeta_{r}(\theta) \Delta \theta
    \label{eq:dw_r}
\end{align}
where
\begin{align}
    \zeta_r(\theta) := 2 (\lambda_r - \frac{1}{\lambda_r^5}) \frac{d \lambda_r}{d \theta} \cdot \pi R^2 t . 
    \label{eq:zeta_r}        
\end{align}

\subsubsection{Virtual Work by Grasping Force}
Let $\Delta \bm{\theta}$ denote a vector storing the virtual displacements of joint angles, defined as
\begin{align}
    \Delta \bm{\theta}
    =
    \begin{bmatrix}
        \Delta \theta_1 & \cdots & \Delta \theta_{n_s}
    \end{bmatrix}^T .
\end{align}
Then, the relationship between the virtual displacement of the finger tip and $\Delta \bm{\theta}$ is obtained as
\begin{align}
    \begin{bmatrix}
        \Delta x \\ \Delta y
    \end{bmatrix}
    =
    \begin{bmatrix}
        \bm{J}_x \\ \bm{J}_y
    \end{bmatrix}
    \Delta \bm{\theta}
    \label{eq:jacobian}
\end{align}
where $\bm{J}_x$ and $\bm{J}_y$ are the Jacobian matrices.

In general, $\Delta \bm{\theta}$ is not uniquely determined when given a displacement of the total bending angle $\Delta \theta$.
Hence, we represent $\Delta \bm{\theta}$ using a coefficient vector $\bm{w}$ as
\begin{align}
    \Delta \bm{\theta} &= \bm{w} \Delta \theta
    \label{eq:weight}
    \\
    \bm{w} 
    & :=
    \begin{bmatrix}
        w_1 & \cdots & w_{n_s}
    \end{bmatrix}^T .
\end{align}
Note that $w_1, \cdots, w_{n_s}$ satisfy the following equation.
\begin{align}
    \sum_{i=1}^{n_s} w_i = 1
\end{align}
The simplest way to determine these coefficients is to set all coefficients to the same value: $w_i = \frac{1}{n_s}$.
In reality, however, this does not always apply.
In particular, these coefficients are different when the fingers grasping an object.
In \cite{ishibashi2024}, we set the value of $\bm{w}$ based on an FEM result of the detailed design for a more detailed analysis, which is also mentioned in Section \ref{sect:fem_analysis}.

\subsubsection{Total Virtual Work Relationship and Grasping Condition}
As we assumed in the beginning of this section, the contact forces are equally distributed among the fingertips. 
Therefore, $F_y$ is given by
\begin{align}
    F_y = \frac{mg}{n} .
\end{align}
Substituting (\ref{eq:pump_work}), (\ref{eq:dw_f}), (\ref{eq:dw_r}), (\ref{eq:jacobian}) and (\ref{eq:weight}) into (\ref{eq:virtual_work}) and eliminating $\Delta \theta$ yield the following relation among $P, \theta, F_x$ and $m$.
\begin{align}
    P(\theta, m, F_x)
    =
    \frac{1}{n a^2 \pi d} \{ 
    n \mu_{sf} \zeta_f(\theta) + \mu_{sr} \zeta_r(\theta) - 
    ( n F_x \bm{J}_x + mg \bm{J}_y) \bm{w}
    \}
    \label{eq:balance_equation}
\end{align}

When the gripper is not grasping an object, we can derive the following relationship between $P$ and $\theta$, substituting $F_x = F_y = 0$ into (\ref{eq:balance_equation}).
\begin{align}
    P = f(\theta) = \frac{1}{n a^2 \pi d} (n \mu_{sf} \zeta_f(\theta) + \mu_{sr} \zeta_r(\theta))
    \label{eq:pressure_model}
\end{align}

\begin{figure}[t]
    \centering
        \includegraphics[width=0.5\hsize]{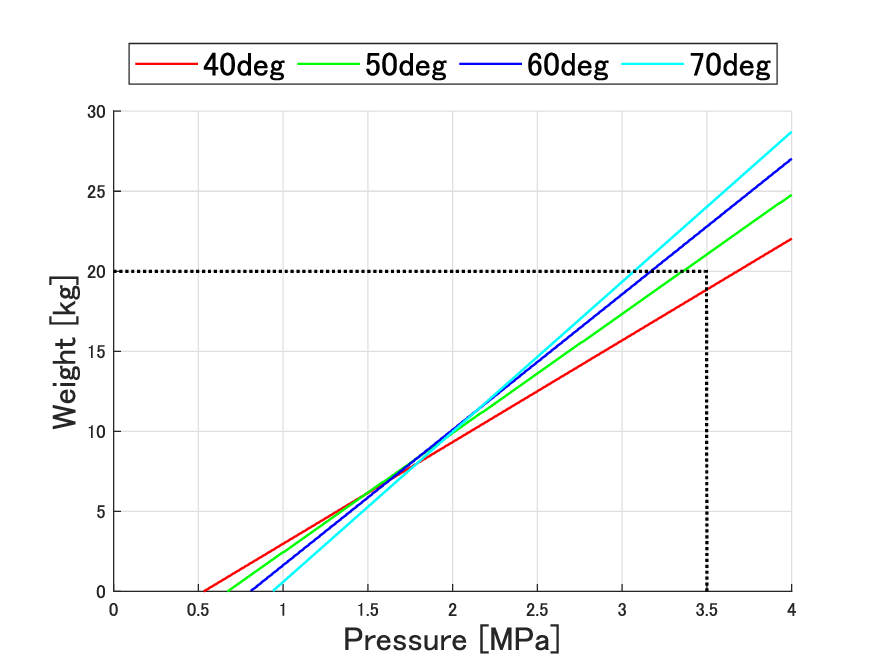}
    \caption{Maximum payload of soft gripper estimated for bending angles of 40$^\circ$, 50$^\circ$, 60$^\circ$, and 70$^\circ$. }
    \label{fig:reservoir_pressure}
\end{figure}

\subsection{Maximum Payload Estimation based on Detailed Parameter Values}

In the presented model, we first determined the values of the parameters $l_0$ and $\theta_0$ to grasp about 20 cm diameter object. 
Because it is hard to systematically optimize the other detailed parameters, we determined those values through cycles of prototyping and checking if the requirements were satisfied, including trial and error. Table \ref{tb:model_param} lists the determined parameters. 

Using the determined parameters, we calculate the maximum payload. 
Grasping an object without slipping requires the following conditions on the friction force:
\begin{align}
    F_v &= F_x \cos(\theta - \theta_0) + F_y \sin(\theta - \theta_0) \\
    F_u &= F_x \sin(\theta - \theta_0) - F_y \cos(\theta - \theta_0) \\
    &-\mu_f F_v \leq F_u \leq \mu_f F_v
    \label{eq:possibility}
\end{align}
where $F_u$ and $F_v$ are the contact forces w.r.t. the $uv$ coordinates in Fig. \ref{fig:finger_model}, and $\mu_f$ is the maximum static friction force coefficient.

Fig. \ref{fig:reservoir_pressure} (a) shows the maximum payload calculated for bending angles of 40$^\circ$, 50$^\circ$, 60$^\circ$, and 70$^\circ$. 
It is seen that the maximum payload can be more than 20 kg when the pressure is 3.5 MPa.

\section{Development of Hydraulic Soft Gripper}
\label{chap:soft_hand}

\subsection{Detailed Structure of Hydraulically-driven Soft Gripper \cite{ishibashi2024}}

\begin{figure}[t]
    \centering
        \includegraphics[width=0.55\hsize]{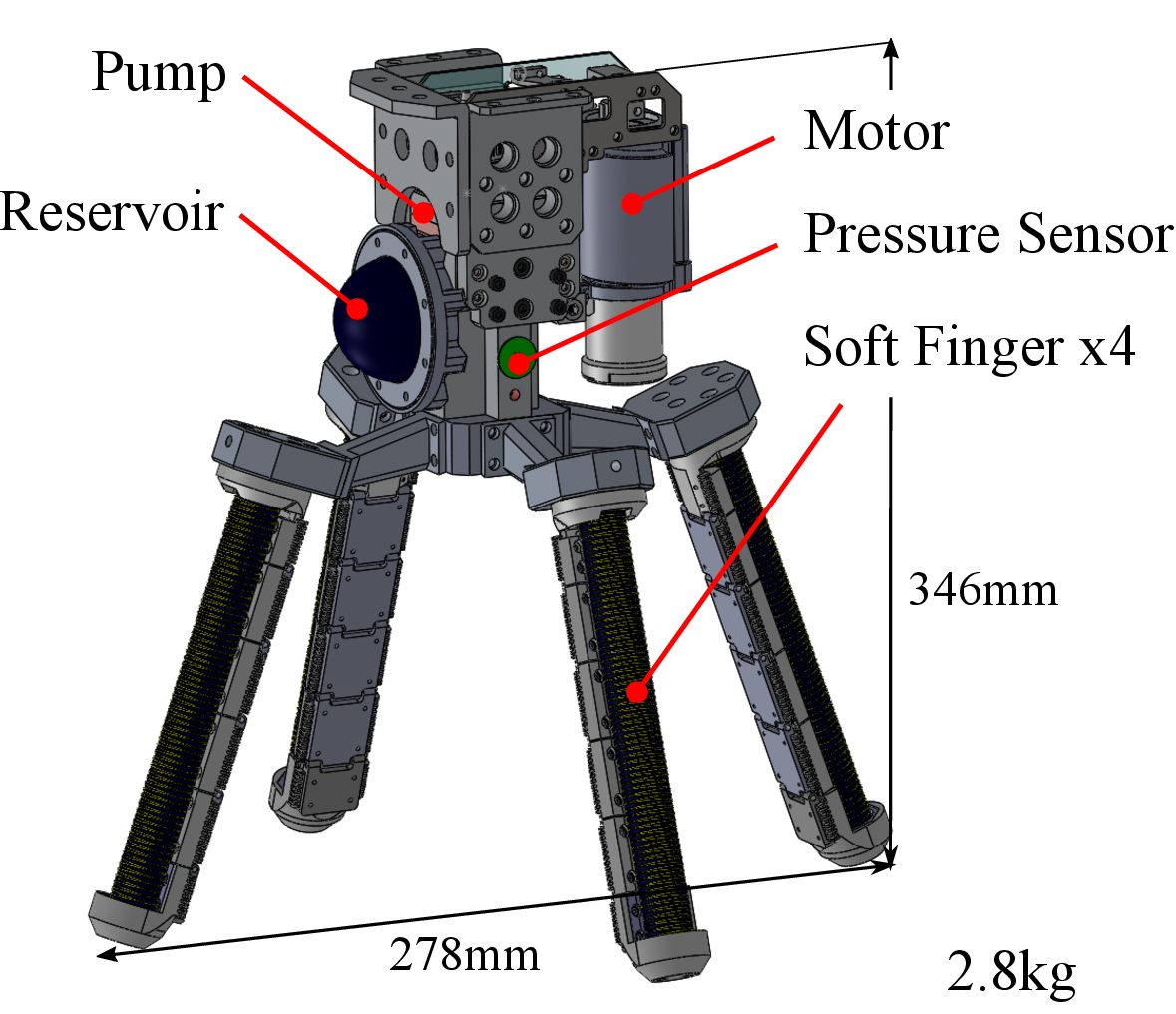}
        \includegraphics[width=0.32\hsize]{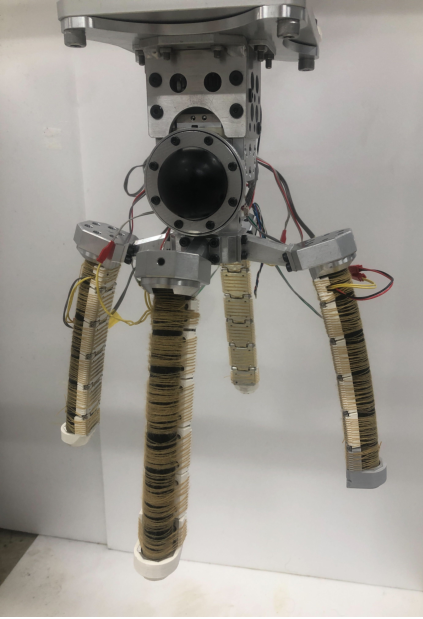}
    \caption{Four-finger hydraulic soft robot gripper. The gripper is composed of soft fingers, pressure sensors, a reservoir for storing oil, a hydraulic pump, and a motor for driving the pump.}
    \label{fig:soft_hand}
\end{figure}

Fig. \ref{fig:soft_hand} shows the developed soft gripper. The width and height are 278 mm and 346 mm, respectively, with a total weight of approximately 2.8 kg, including the actuation unit (1.67 kg) consisting of a hydraulic pump and electric motor. 
We used a hydraulic pump (dimensions: 42 x 47.2 x 22.8 mm; weight: 228 g) developed in \cite{Komagata2019}.
Two pressure sensors are equipped on the high- and low-pressure sides of the pump flow path, which enables us to measure the differential pressure for the pressure control by the motor current.
\begin{figure}[t]
    \centering
    \includegraphics[width=0.7\hsize]{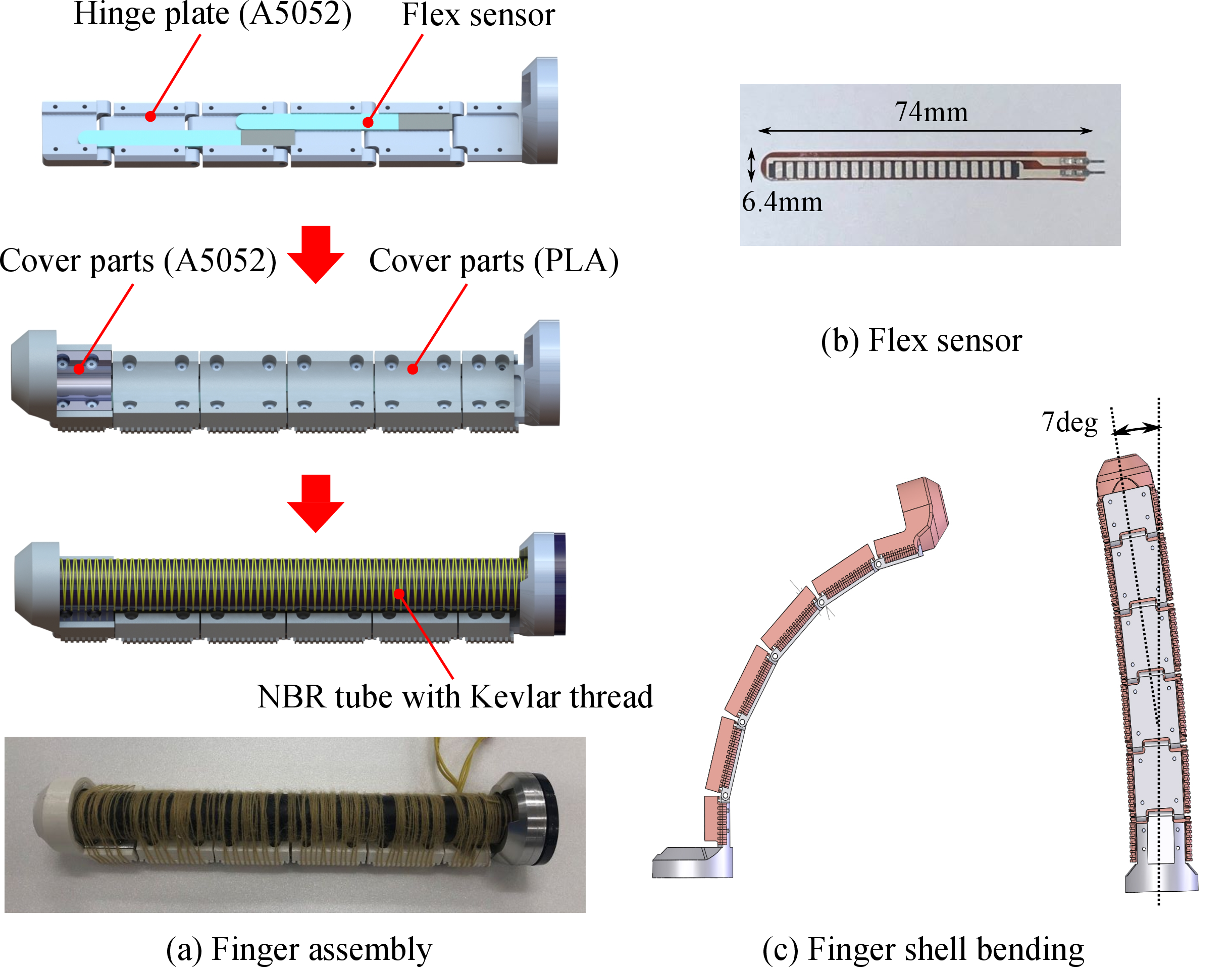}
    \caption{Fabricated soft finger. (a) Structure of soft finger, (b) Resistive flexible sensor, (c) Image of finger shell bending.}
    \label{fig:finger_detail}
\end{figure}

Fig. \ref{fig:finger_detail} shows the structure of the developed soft finger. The fabrication process is summarized as follows:
\begin{enumerate}
    \item The exoskeleton parts of the fingers are made of A5052 to prevent excessive twisting in the axial direction. These parts are connected by hinge joints. Two flexible sensors (SPECTRASYMBOL-006, Spectra Symbol, Inc) shown in Fig. \ref{fig:finger_detail} (b) are inside the exoskeleton to measure the bending angle.
    \item Next, a 3D-printed cover is attached to each exoskeleton part. The covers are made of PLA, except for the fingertip, which is made of A5052 based on an FEM analysis discussed in the next subsection.
    \item Finally, an NBR tube ($\phi 18 \times$ 180 mm, and 4 mm thick) constrained by Kevlar thread is inserted into the exoskeleton parts.
\end{enumerate}

\subsection{FEM Analysis\label{sect:fem_analysis}}

Based on the detailed design of the developed gripper, we conducted an FEM analysis to select suitable materials for the necessary parts, and also determine the specific values of $\bm{w}$ in the mathematical model. 
Using the parameters listed in Table \ref{tb:fem_param}, we simulated a finger movement setting the driving pressure to 2.0 MPa, 3.0 MPa and 4.0 MPa.

\begin{figure}[t]
    \centering
    \includegraphics[width=1.0\hsize]{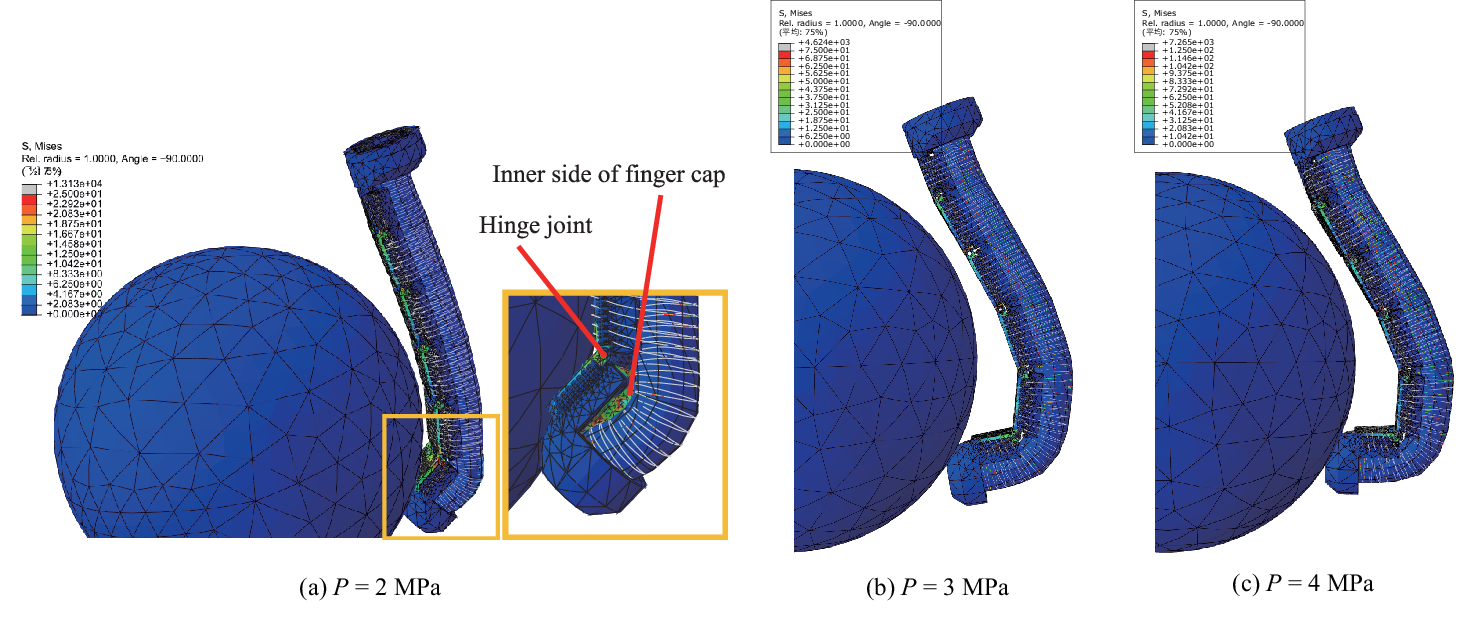}
    \caption{FEM simulations of an object grasping under pressure of 2.0 MPa, 3.0 MPa and 4.0 MPa. This figure indicates that stress is concentrated at the cap and hinge points at the tips of the fingers.}
    \label{fig:abaqus}
\end{figure}

Fig. \ref{fig:abaqus} shows the results of the FEM analysis, and Table \ref{tb:stress} summarizes the maximum stress in the fingertip components. 
When applying 4.0 MPa pressure, the maximum stress was 235.4 MPa at the fingertip cap and 170.3 MPa at the hinge joint in the fingertip. Since the fracture strength of PLA is 55 MPa, we selected A5052 for these parts due to its higher strength. 
Moreover, the strength of NBR with 70 Shore A hardness is 12.7 MPa, which affords to the driving pressure.

\begin{table}[t]
  \centering
    \footnotesize
    \caption{Parameter used in FEM simulation}
    \label{tb:fem_param}
      \begin{tabular}{c|c} \hline
        Mooney-Rivlin coefficient of NBR tube  & $C_{10}=0.41$, $C_{01}=0.43$ \rule[0pt]{0pt}{15pt} \\
        Young's modulus of A5052 metal parts & 68 MPa \rule[0pt]{0pt}{15pt} \\
        Poisson's ratio of A5052 metal parts & 0.33 \rule[0pt]{0pt}{15pt} \\
        Young's modulus of PLA parts & 3.5 MPa \rule[0pt]{0pt}{15pt} \\
        Poisson's ratio of PLA parts & 0.3 \rule[0pt]{0pt}{15pt} \\
        Young's modulus of Kevlar thread beam & 31 MPa \rule[0pt]{0pt}{15pt} \\
        Poisson's ratio of Kevlar thread beam & 0.36 \rule[0pt]{0pt}{15pt} \\
        Radius of Kevlar thread beam & 0.2 mm \rule[0pt]{0pt}{15pt} \\ \hline
      \end{tabular}
\end{table}

\begin{table}[t]
  \centering
    \footnotesize
    \caption{Comparison of maximum stress and material fracture strength applied to fingers.}
    \label{tb:stress}
      \begin{tabular}{c|c|c|c}
        Pressure applied to the finger & 2.0MPa & 3.0MPa & 4.0MPa \rule[0pt]{0pt}{15pt} \\
        \hline
        Maximum stress of a cap part of the finger & 124.8MPa & 153.4MPa & 235.4MPa \rule[0pt]{0pt}{15pt} \\
        Maximum stress of a hinge structure & 82.3MPa & 114.4MPa & 170.3MPa \rule[0pt]{0pt}{15pt} \\
        Maximum stress of a rubber tube & 3.3MPa & 3.9MPa & 4.9MPa \rule[0pt]{0pt}{15pt} \\
        \hline \hline
        PLA strength & \multicolumn{3}{|c}{55MPa} \rule[0pt]{0pt}{15pt} \\
        A5052 strength & \multicolumn{3}{|c}{260MPa} \rule[0pt]{0pt}{15pt} \\
        NBR strength & \multicolumn{3}{|c}{12.7MPa} \rule[0pt]{0pt}{15pt} \\
        \hline
      \end{tabular}
\end{table}

It is observed from Fig. \ref{fig:abaqus} that the 5th joint bent more largely than the other joints when the pressure is increased.
We consider that this is because the fingertip contacts with the object on a fixed point.
This result implies that the assumption in the previous section does not apply --- all joints do not have the same angle but different angles.
Therefore, we determine the value of the coefficient vector $\bm{w}$ based on the FEM result in \cite{ishibashi2024}.
In the FEM simulation, the values of $\varDelta \theta_i \ (i=1, \cdots, 5)$ were calculated in each step of the simulation, and we used those values to determine $\bm{w}$.
As a result, the value of $\bm{w}$ was set as $\bm{w} = (-0.2, 0.0, 0.1, 0.3, 0.8)^T$.
Moreover, the modeling using this value of $\bm{w}$ was validated in \cite{ishibashi2024} based on a preliminary experiment on a 6 kg object, which will be shown in Section \ref{chap:evaluation}.

\section{Experimental Validations}
\label{chap:evaluation}
\begin{figure}[t]
        \centering
	 	\includegraphics[width=0.3\hsize]{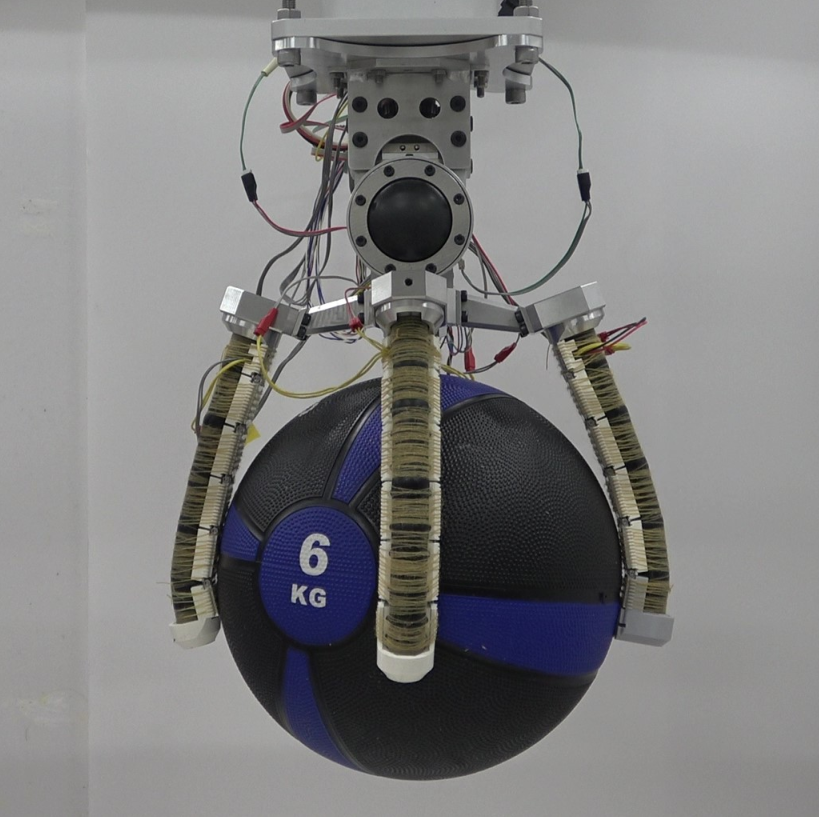}
	\caption{Pre-experiment to re-estimate the maximum payload. In this experiment, we tested a grasping of a 6 kg dumbbell.\label{fig:angle_force}}
\end{figure}

\begin{figure}[t]
    \centering
    \includegraphics[width=0.5\hsize]{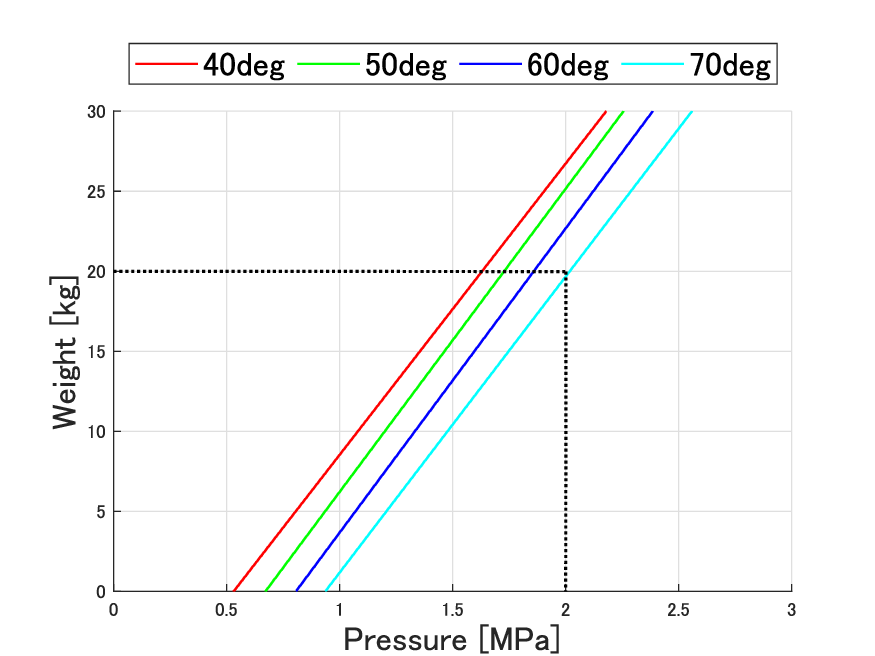}
    \caption{Maximum payload of four-finger soft gripper when wrapping an object. For a pressure of 2.0 MPa, the gripper can grasp a maximum weight of 27 kg at 40 degrees and a maximum weight of 20 kg at 70 degrees.}
    \label{fig:max_weight_20kg}
\end{figure}
\begin{figure}[t]
    \centering
    \includegraphics[width=0.4\hsize]{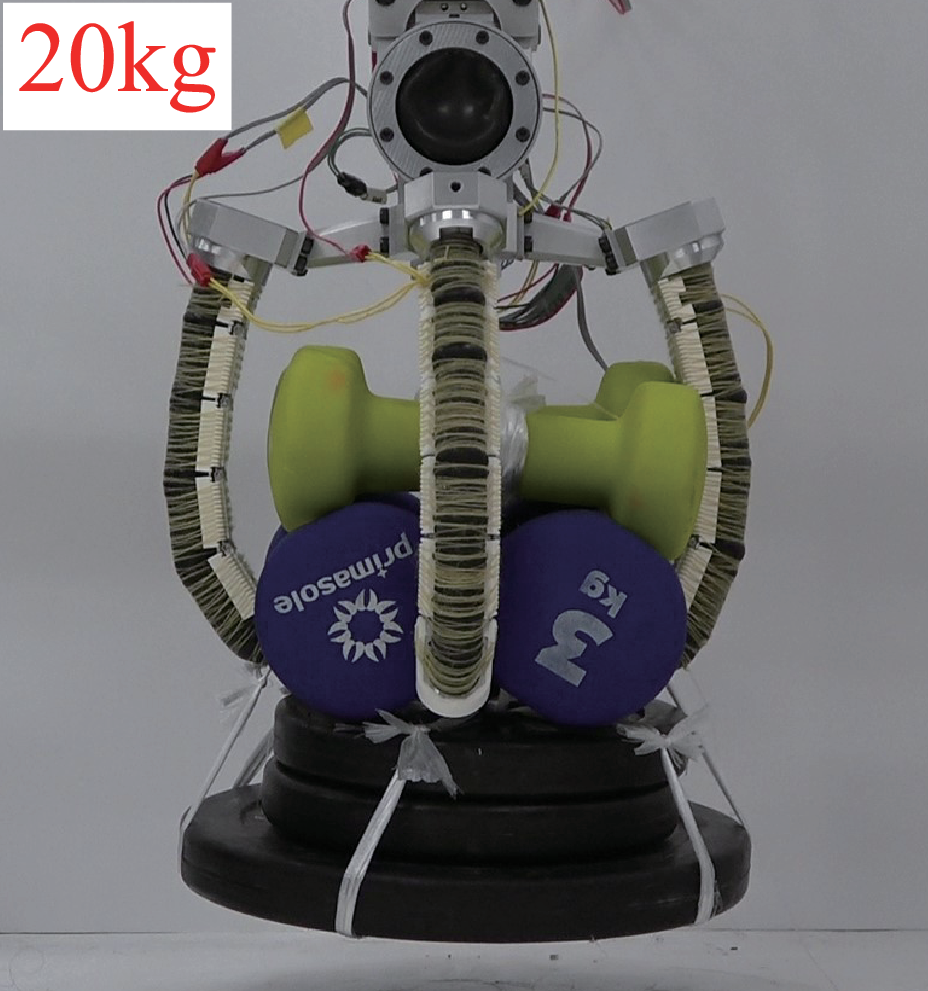}
    \caption{Experimental scene of 20 kg payload test. In this experiment, a four-finger soft gripper grasped a dumbbells whose total weight is 20 kg and held it for 10 s by the 2.0 MPa pressure control.}
    \label{fig:20kg_grasp}
\end{figure}
\subsection{Maximum Payload}

We experimentally validate the maximum payload of the developed gripper. 
Before the experiment, we re-estimate the maximum payload, which was originally presented in Fig. \ref{fig:reservoir_pressure}, based on the mathematical model using the value of $\bm{w}$ determined as mentioned in the previous section. 
The relationship between the finger angle and the grasping force using this value was validated also in \cite{ishibashi2024} with a preliminary experiment using the 6 kg object (shown in Fig. \ref{fig:angle_force}).
Fig. \ref{fig:max_weight_20kg} shows the updated result of the maximum payload to a given finger bending angle. 
Given a pressure of 2.0 MPa, the maximum payload is estimated as 27 kg at 40$^\circ$ and 20 kg at 70$^\circ$. 

Fig. \ref{fig:20kg_grasp} shows an experimental scene in which the developed soft gripper grasps a set of dumbbells whose total weight is 20 kg and holds it for 10 s.
The results of the 20 kg payload is consistent with the result of Fig. \ref{fig:max_weight_20kg}. 

Although we set the value of $\bm{w}$ based on the FEM result with a 240 mm diameter object, this condition does not always apply to unknown shape object.
Indeed, the shape of the 20 kg object was largely different from the 240 mm diameter sphere object.
Rather, this result shows that the payload estimation based on the developed model can be applied to such a different condition as the softness of the developed hand absorbed an uncertainty.

\subsection{Examples of Grasping Various Objects}
\begin{figure}[t]
    \centering
    \includegraphics[width=1.0\hsize]{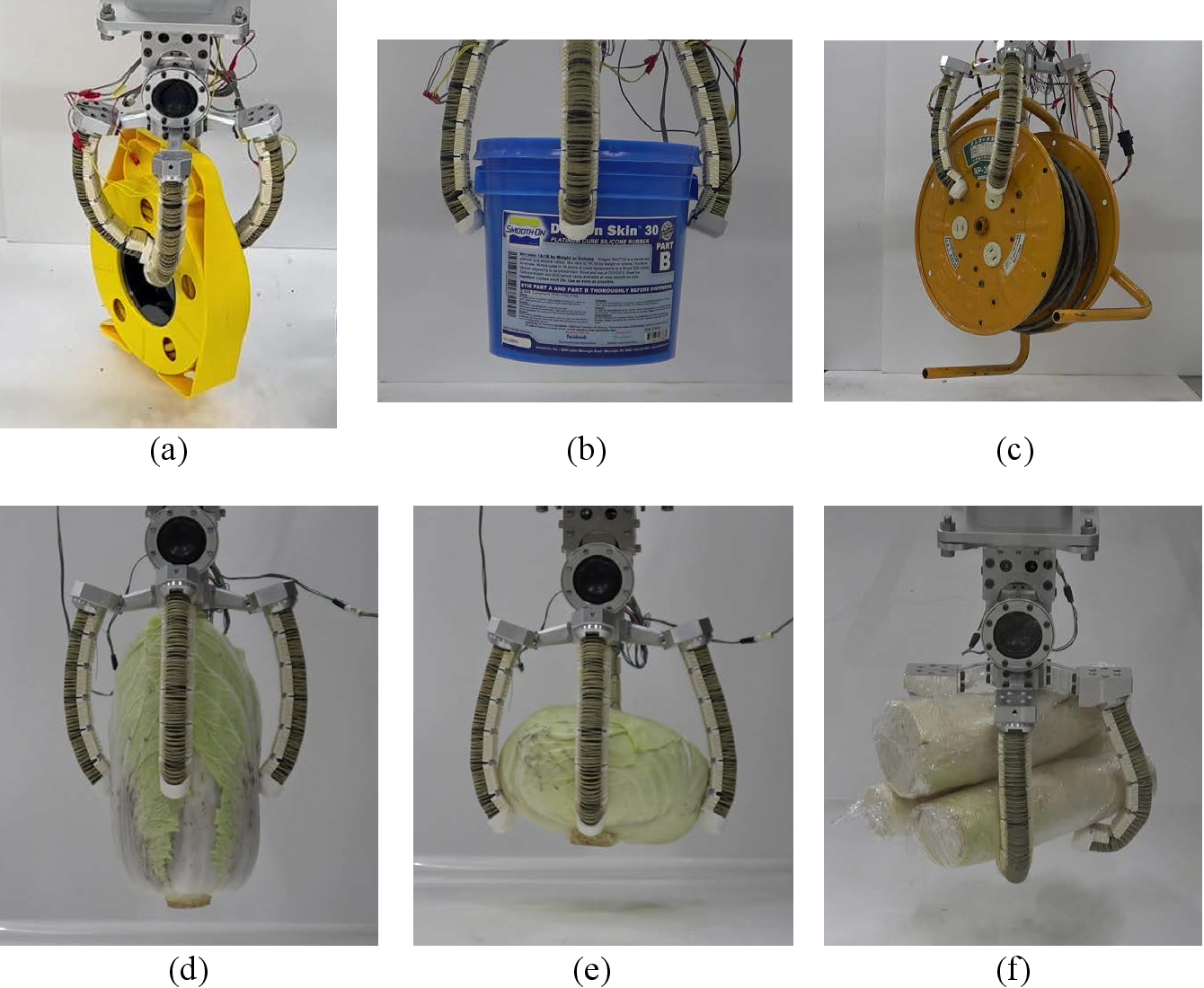}
    \caption{Grasping various objects: (a) cable reel (5.8 kg), (b) bucket (4 kg), (c) cable real (6 kg), (d) Chinese cabbage (2.6 kg), (e) cabbage (2.3 kg), and (f) three radishes ($\simeq$ 2.1 kg).}
    \label{fig:grasping_vegetables}
\end{figure}

Moreover, we validated the adaptability of the developed soft gripper to various objects as shown in Fig. \ref{fig:grasping_vegetables}, including some large size vegetables.
The simple pressure control enabled the soft gripper to grasp various industrial products including Fig. \ref{fig:grasping_vegetables}(a)(c) cable reals and (b) a bucket.
One of possible applications of this gripper is a picking task in a vegetable factory.
We also demonstrated (d) Chinese cabbage, (e) cabbage and (f) radishes were grasped by the developed gripper.
These results show that the developed hand achieved not only high payload but also high adaptability to variety in different conditions of shape and surface.

\subsection{Closed-loop Control of Finger Bending Angle}
\label{chap:control}

\begin{figure}[t]
    \centering
    \includegraphics[width=0.7\hsize]{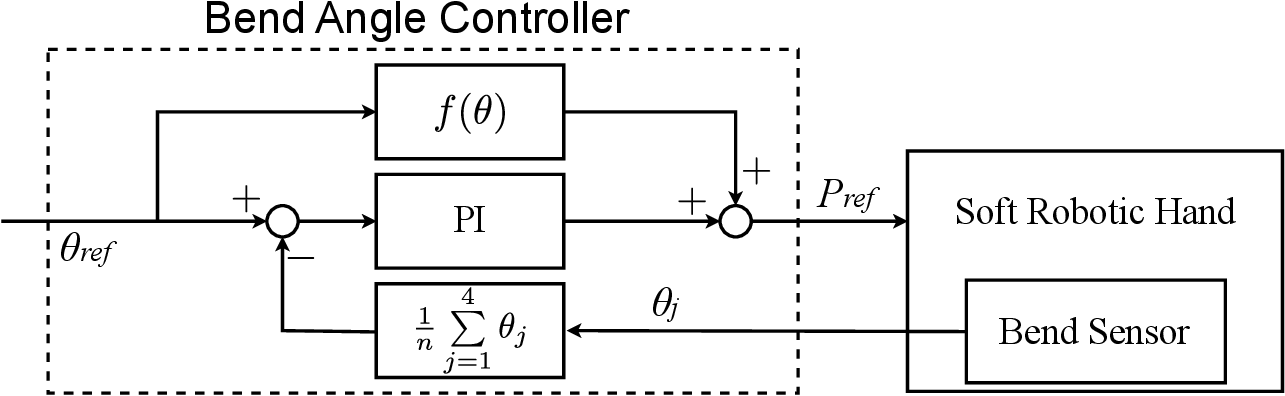}
    \caption{Two-degree-of-freedom control of the bending angle of the fingers using the relationship between $\theta$ and $P$ obtained by (\ref{eq:pressure_model}). The average angle of fingers was controlled to follow the commanded angle since each finger is connected to the same flow path.}
    \label{fig:block_diagram}
\end{figure}

Finally, we validated a feedback control of the finger angles. 
While we demonstrated a robustness of the feedback control to horizontal-direction disturbance in \cite{ishibashi2024}, we demonstrate a robustness to vertical-direction disturbance in this study.

Fig. \ref{fig:block_diagram} shows a block diagram of the 2-DOF control that we implemented in \cite{ishibashi2024}. The reference value of the pressure is given by
\begin{align}
    \label{eq:finger_control}
        P_{ref} &= f(\theta_{ref}) + K_P \theta_e + K_I \int{\theta_e dt}
        \\
    \theta_e &:= \theta_{ref} - \frac{1}{n} \sum_{j=1}^{n} \theta_j
    \label{eq:finger_error}
\end{align}
where $f(\theta_{ref})$ is the feedforward term obtained by \eqref{eq:pressure_model}, and $K_P$ and $K_I$ are PI gains.
As defined in \eqref{eq:finger_error}, we fed back the error between the reference angle $\theta_{ref}$ and the average angle of all fingers.

\begin{figure}[t]
    \centering
    \includegraphics[width=0.7\hsize]{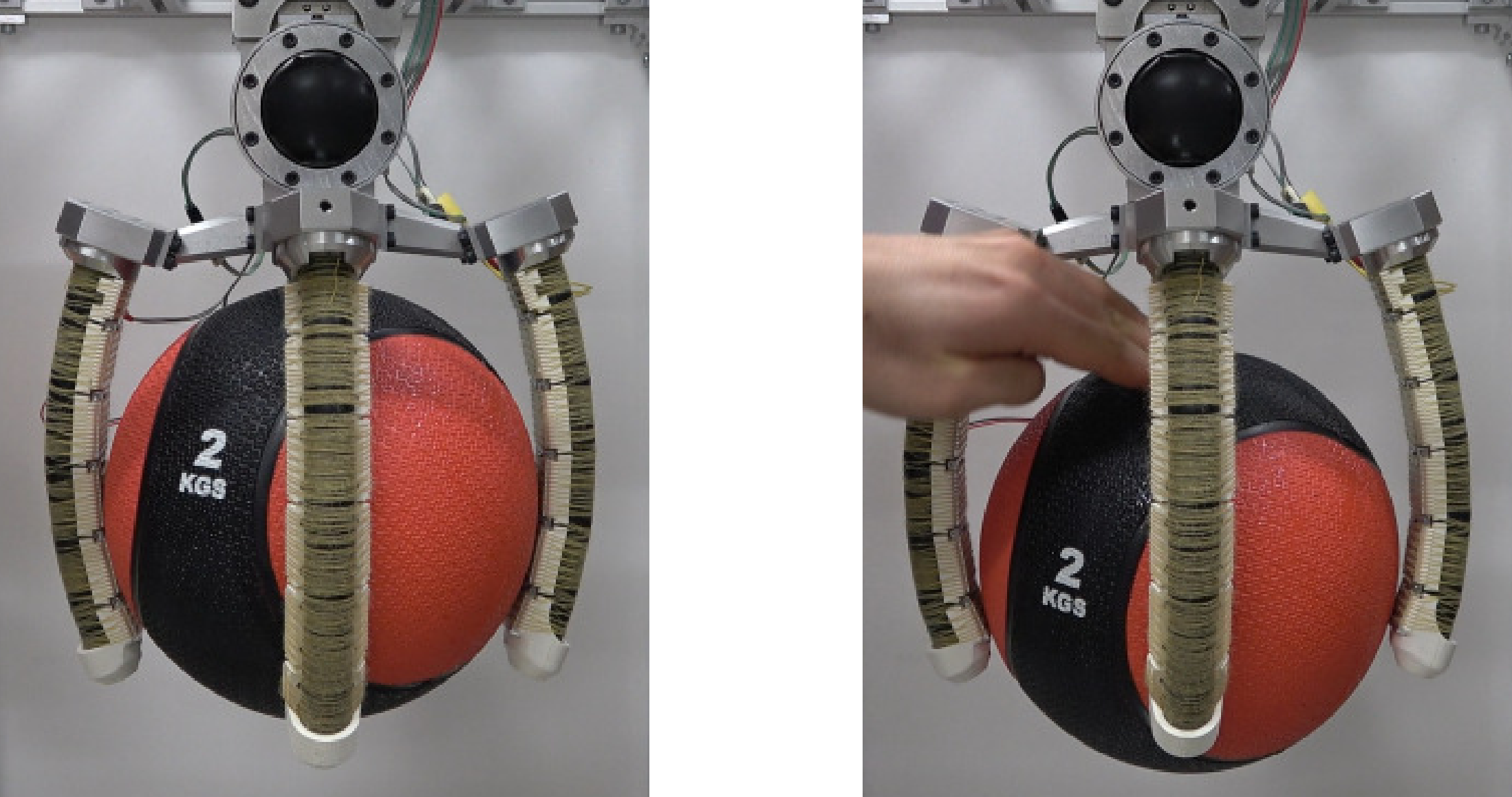}
    \caption{Experimental scenes where a disturbance was added manually in the vertical direction while the gripper was holding the 2 kg medicine ball.}
    \label{fig:disturbance_picture}
\end{figure}

\begin{figure}[t]
    \centering
    \includegraphics[width=0.7\hsize]{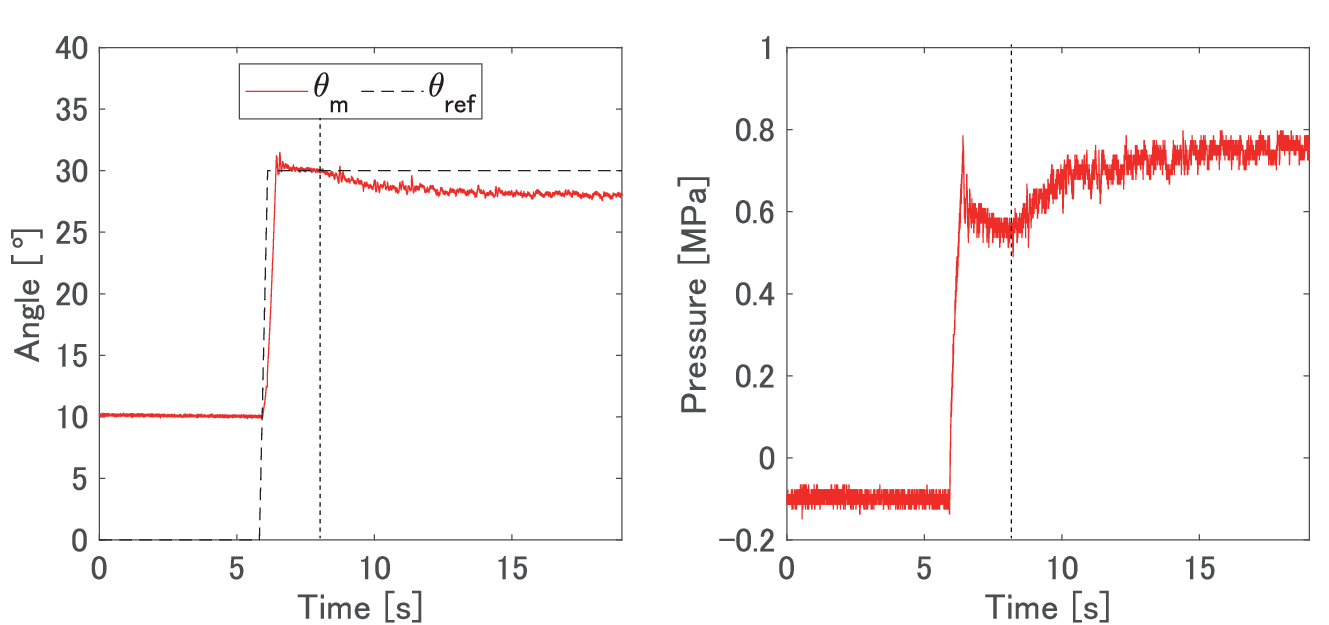}
    \caption{Values of the finger angle and pressure in the experiment shown in Fig. \ref{fig:disturbance_picture}.}
    \label{fig:bend_dp}
\end{figure}

Fig. \ref{fig:disturbance_picture} shows experimental scenes where a disturbance was added manually in the vertical direction while the gripper was holding the 2 kg medicine ball.
Fig. \ref{fig:bend_dp} shows the values of the finger angle and the pressure in the experiment, in which the vertical dashed line indicates the moment when the downward disturbance was added.
It is observed that higher pressure was generated by the pump to keep holding the ball.

\section{Discussions\label{sect:discussion}}
\subsection{Payload-to-weight Ratio}
\begin{table}[t]
    \footnotesize
    \centering
    \begin{tabular}{l||ccc}
         & Self weight [g] & Payload-to-weight ratio & Type \\ \hline
        (A) \cite{Hong2023} & 0.4 & 16000 & Kirigami \\
        (B) \cite{Cacucciolo2019} & 1.5 & 1086 & Electro-adhesion \\
        (C) \cite{Li2019} & 97 & 123 & Pneumatics (negative pressure) \\
        (D) \cite{Yap2016} & 277 & 18 & Pneumatics (positive pressure) \\
        Ours & 1130 \footnote{Excluding the actuation unit.} & 17.7 & Hydraulics (positive pressure) \\
    \end{tabular}
    \caption{Comparison of payload-to-weight ratio with other soft robot grippers.}
    \label{tab:payload_to_weight_raito_comparison}
\end{table}

\begin{figure}[t]
    \centering
    \includegraphics[width=0.5\hsize]{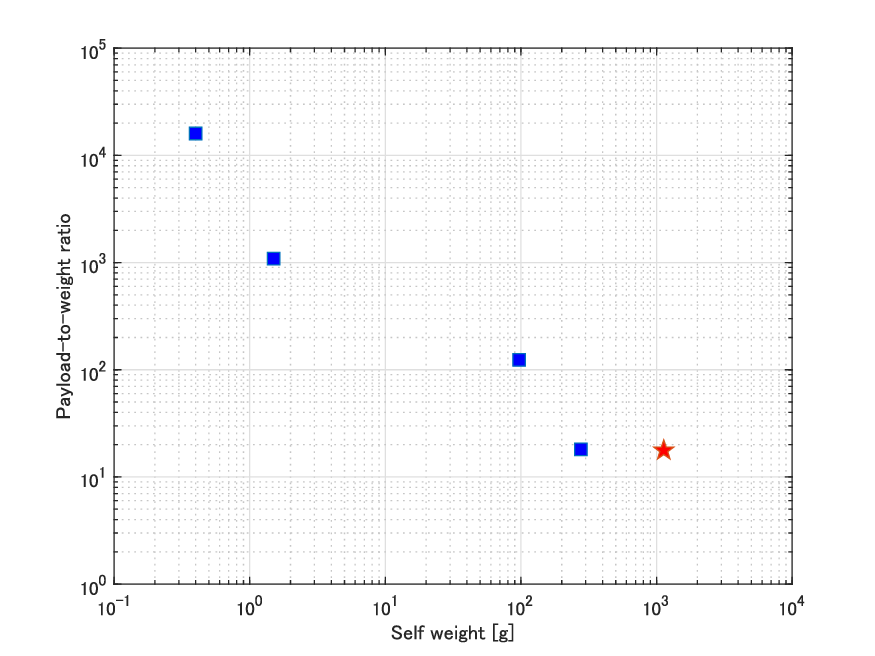}
    \caption{Payload-to-weight ratio versus the mass of soft grippers compared in Table \ref{tab:payload_to_weight_raito_comparison}}
    \label{fig:payload_to_weight_ratio_summary}
\end{figure}

The weight of the developed soft gripper excluded the actuation unit is 1.13 kg. 
As an actuation source is usually built outside in most soft grippers, we calculate the payload-to-weight ratio as 17.7 ($\simeq 20 / 1.13$) using the weight excluding the actuation unit.
We compare this value with other soft grippers that have high payload-to-weight ratio.
Table \ref{tab:payload_to_weight_raito_comparison} shows the comparison of the payload-to-weight ratio, including the actuation type of each soft gripper, and Fig.\ref{fig:payload_to_weight_ratio_summary} shows a plot of the payload-to-weight ratio versus the weight of soft grippers.
Note that the vertical and horizontal axes of Fig.\ref{fig:payload_to_weight_ratio_summary} are log scales.
Soft grippers using Kirigami structure \cite{Hong2023} or electro-adhesion \cite{Cacucciolo2019} achieved high payload-to-weight-ratio (16000 in \cite{Hong2023} and 1086 in \cite{Cacucciolo2019}) with light self-weight less than a few grams.
However, the payload-to-weight ratio tends to decrease as the self weight increases.
This implies that the payload-to-weight ratio does not always scale to larger weight.
The payload-to-weight ratio of our soft gripper is as large as that of Yap et al. \cite{Yap2016}, but slightly larger than a line approximated by least-square of the other soft grippers \cite{Hong2023,Cacucciolo2019,Li2019,Yap2016}.
Rather, we consider that an advantage of the developed gripper compared to these studies is the capacity of a large size object with 20 -- 30 cm diameter.

\begin{figure}[t]
        \centering
		\subfigure[2 kg]{
			\includegraphics[width=0.45\hsize]{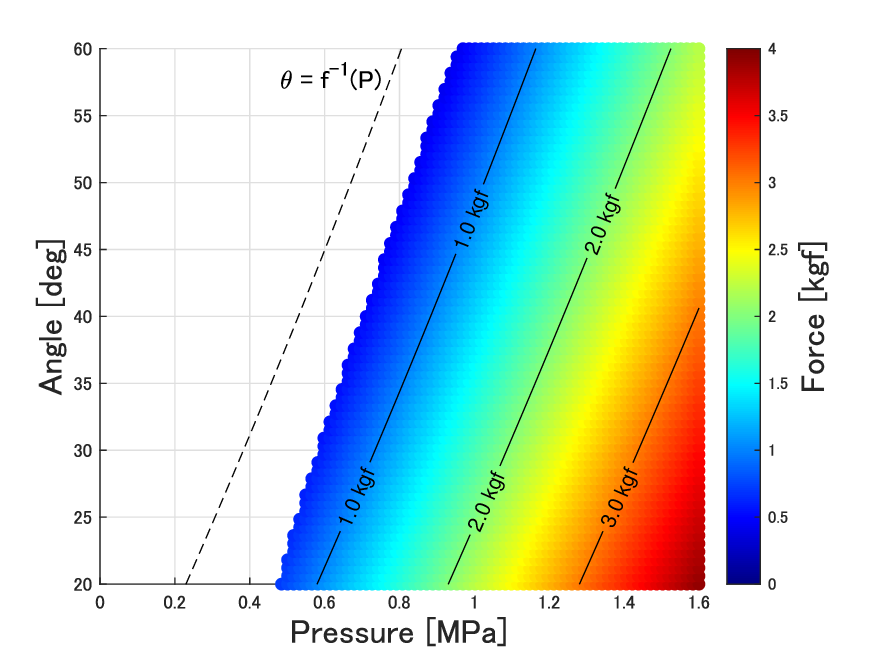}}
		\subfigure[6 kg]{
	 		\includegraphics[width=0.45\hsize]{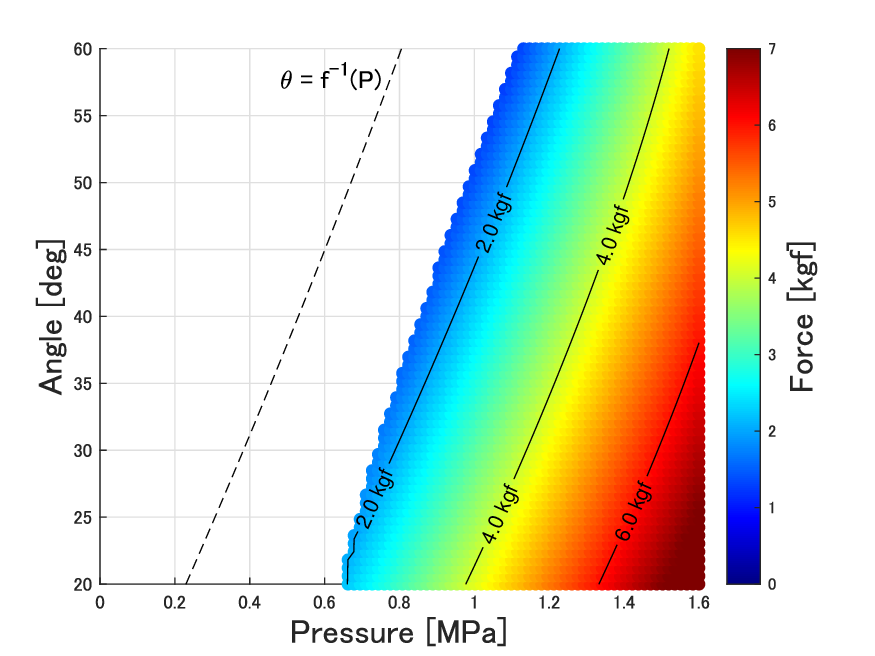}}
		\subfigure[10 kg]{
			\includegraphics[width=0.45\hsize]{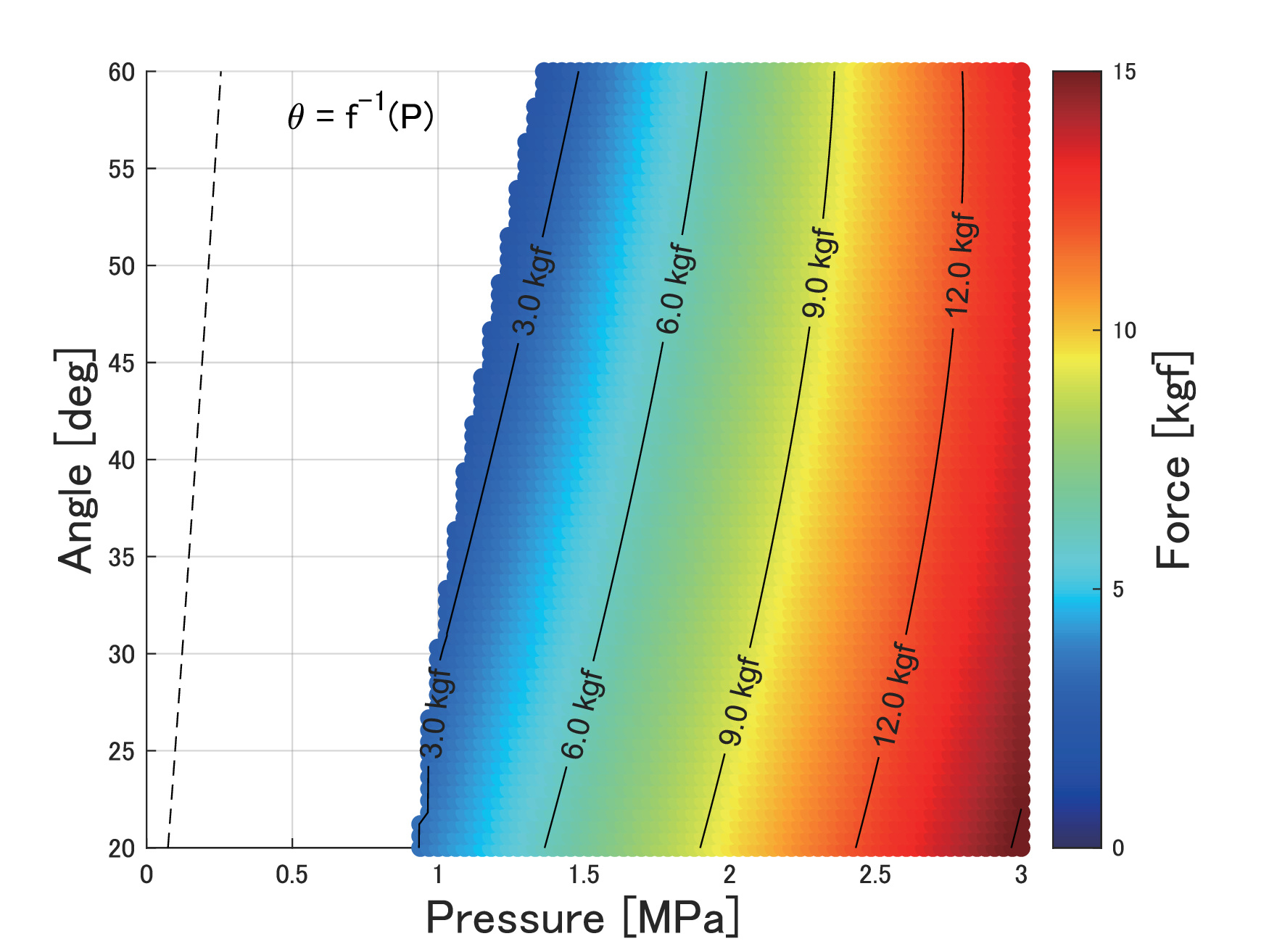}}
		\subfigure[20 kg]{
			\includegraphics[width=0.45\hsize]{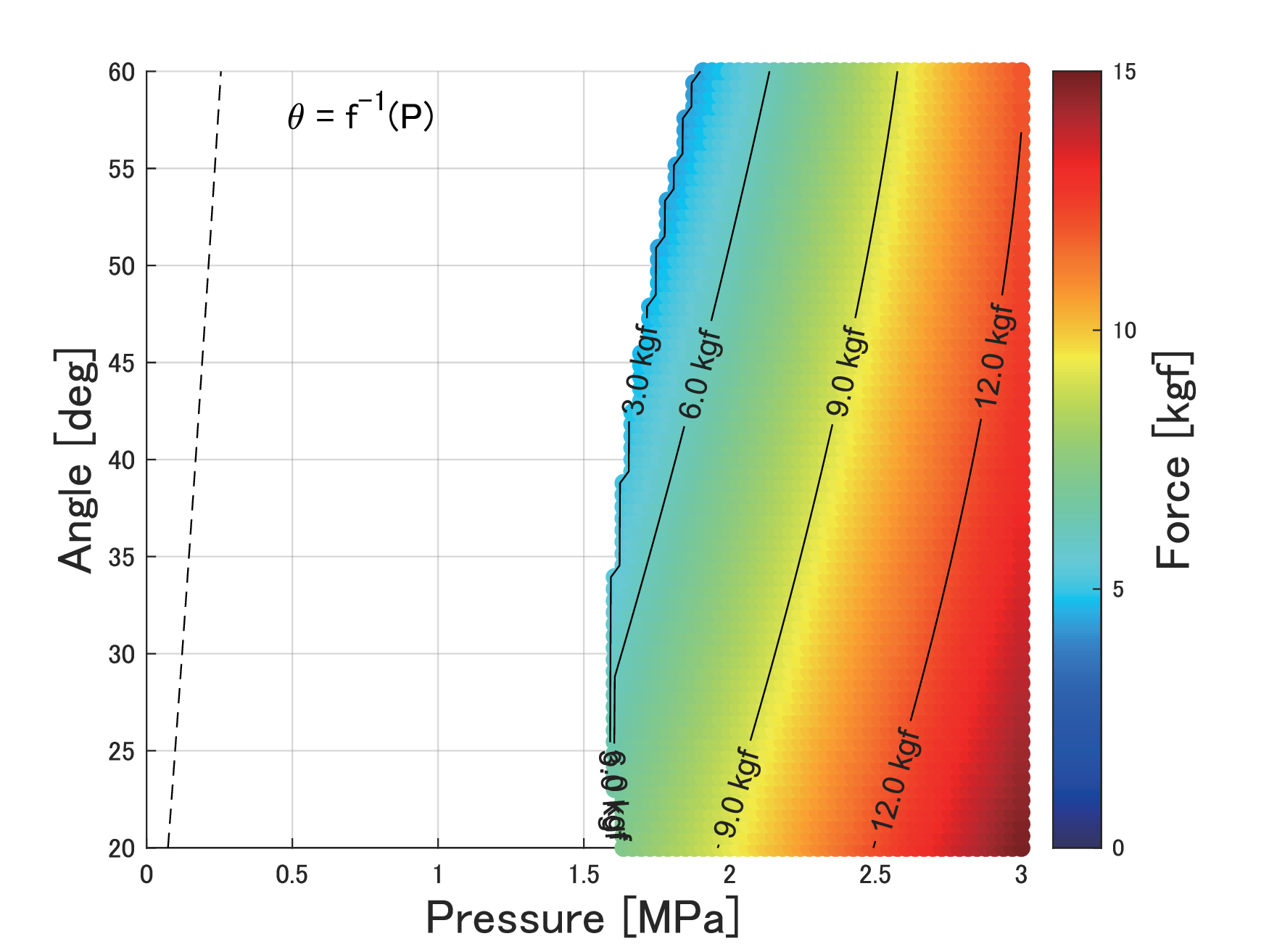}}
		\caption{Relationship between the pressure and finger angle for different conditions of the object weight. Object are graspable in the colored area. The difference of color indicates the grasping force. The dotted line indicates the case without grasping an object, represented by (\ref{eq:pressure_model}).
  \label{fig:grasp_possibility}}
\end{figure}

\subsection{Calculation of Graspable Area}
Furthermore, the obtained model can be used to visualize the conditions under which the gripper can grasp an object. The conditions for grasping an object were represented by (\ref{eq:possibility}). Fig. \ref{fig:grasp_possibility} shows the graspable area of an object when the gripper wraps an object. The vertical axis represents the driving pressure, the horizontal axis represents the bending angle, and the color difference indicates the magnitude of the grasping force $F_v$. This relationship allows for calculating the pressure $P$ required to grasp an object and the grasping force $F_v$ when the approximate weight and size of the object are known by determining $\theta$. 

\section{Conclusion}
\label{chap:conclusion}

In this study, we developed a hydraulically-driven soft gripper and presented the design methodology for achieving the 20 kg payload. 
The key results are summarized as follows:

\begin{enumerate}
    \item We selected appropriate material based on the result of the FEM analysis.
        When applying 4.0 MPa pressure, the maximum stress was 235.4 MPa at the fingertip cap and 170.3 MPa at the hinge joint in the fingertip. Since the fracture strength of PLA is 55 MPa, we selected A5052 for these parts due to its higher strength.
    \item We estimated the maximum payload based on the previously developed model and the friction condition, and confirmed the maximum payload would be more than 20 kg.
    In the original estimation, we assumed the coefficients used in the model had the same value.
    However, the FEM analysis showed that the joint angle were different when the gripper grasped an object.
    Therefore, we determined the coefficient values from the FEM result, which was validated with the preliminary experiment with the 6 kg object, and re-calculated the payload, which showed that the maximum payload was 20 kg when the total finger angle was 70$^\circ$.
    \item The maximum payload was experimentally validated.
    We also experimentally demonstrated the adaptability of the developed soft gripper to various objects with different shape and size.
    Moreover, the closed-loop control of the finger angle demonstrated the robustness to vertical disturbances.
    These results imply that the developed soft gripper achieved the 20 kg payload without loosing the flexibility, adaptability and robustness.
\end{enumerate}

\section*{Acknowledgement}
This research was supported by the National Agriculture, Forestry and Fisheries Research Organization's International Competitiveness Enhancement Technology Development Project "Development of Agricultural Work Automation Technology Utilizing Robot Arms Suitable for Agricultural Products" (Research Leader: Takanori Fukao).

\bibliographystyle{bib/tfnlm}
\bibliography{bib/reference}

\end{document}